%% file: main.tex
\documentclass{article}

\input{setup/packages}


\input{setup/metadata}

\begin{document}
\maketitle

\begin{abstract}
\input{sections/00_abstract}

\end{abstract}


\input{sections/01_introduction}
\input{sections/02_related_work}
\input{sections/03_method}
\input{sections/04_experiments}
\input{sections/05_conclusion}


\input{sections/99_references}

\newpage
\appendix

\input{appendices/00_hardware_workflow}
\input{appendices/01_kalmanfilter}
\input{appendices/02_task_criteria}
\input{appendices/03_reward_function}
\input{appendices/04_kalman_config}

\input{appendices/05_domain_randomization}
\input{appendices/06_train_config}
\end{document}

%% file: setup/packages.tex
\newif\ifarxiv
\arxivtrue 

\ifarxiv
  \usepackage[preprint]{corl_2026}
\else
  \usepackage[final]{corl_2026}
\fi

\usepackage{graphicx}    
\usepackage{amsmath}     
\usepackage{amssymb}     
\usepackage{booktabs}    
\usepackage{hyperref}    
\usepackage{cleveref}    
\usepackage{color}       

\usepackage{amsmath}  
\usepackage{amssymb}  
\usepackage{booktabs}
\usepackage{multirow}
\usepackage{array}
\usepackage{tabularx}
\usepackage{caption}
\usepackage{subcaption}
\usepackage{xcolor}
\usepackage{float}
\usepackage{placeins}
\usepackage{hyperref}
\usepackage{adjustbox}
\usepackage{arydshln}
\usepackage{dsfont}

\graphicspath{{figures/}}

%% file: setup/metadata.tex
\title{SigLoMa: Learning Open-World Quadrupedal Loco-Manipulation from Ego-Centric Vision}

\author{
  Shiyi Chen$^1$, Haiyi Liu$^1$, Mingye Yang$^2$, Jiaqi Zhang$^1$, Debing Zhang$^{1,*}$ \\
  $^1$Tsinghua University \qquad $^2$Imperial College London \\
  \url{https://11chens.github.io/SigLoMa/}
}

\ifarxiv
\makeatletter
\renewcommand{\@noticestring}{\vspace{-1em}$^*$Corresponding author}%
\makeatother
\fi

\newcommand{\PaperKeywords}{Quadrupedal Loco-manipulation, Ego-centric Vision, Sim-to-Real}

%% file: sections/00_abstract.tex
Designing an open-world quadrupedal loco-manipulation system is highly challenging. Traditional reinforcement learning frameworks utilizing exteroception often suffer from extreme sample inefficiency and massive sim-to-real gaps. Furthermore, the inherent latency of visual tracking fundamentally conflicts with the high-frequency demands of precise floating-base control. Consequently, existing systems lean heavily on expensive external motion capture and off-board computation. To eliminate these dependencies, we present SigLoMa, a fully onboard, ego-centric vision-based pick-and-place framework. At the core of SigLoMa is the introduction of Sigma Points, a lightweight geometric representation for exteroception that guarantees high scalability and native sim-to-real alignment. To bridge the frequency divide between slow perception and fast control, we design an ego-centric Kalman Filter to provide robust, high-rate state estimation. On the learning front, we alleviate sample inefficiency via an Active Sampling Curriculum guided by Hint Poses, and tackle the robot's structural visual blind spots using temporal encoding coupled with simulated random-walk drift. Real-world experiments validate that, relying solely on a 5Hz (200 ms latency) open-vocabulary detector, SigLoMa successfully executes dynamic loco-manipulation across multiple tasks, achieving performance comparable to expert human teleoperation.

%% file: sections/01_introduction.tex
\section{Introduction}
\label{sec:introduction}

Equipping quadrupedal robots with manipulation capabilities holds immense practical value for unstructured environments. However, precise loco-manipulation coupled with a floating base remains fundamentally challenging. While recent works have explored Reinforcement Learning (RL) methods in simulation to coordinate locomotion~\cite{anymal_parkour, zhuang2023robot, cheng2023parkour, him, MoE-Loco} and manipulation~\cite{fu2023deep, eth_badminton, vbc}, external perception remains a prominent bottleneck. In simulation, rendering dense visual inputs incurs high memory overhead, drastically reduces environment parallelism, and leads to highly inefficient training. Furthermore, deploying such vision-based policies introduces a profound sim-to-real gap~\cite{yu2022visual, agarwal2023legged}. 

Most current perception pipelines rely on dense geometric information (e.g., depth maps~\cite{miki2022learning, eth2025robust} or point clouds~\cite{reasan, omni_perception}) to narrow this gap; however, they inherently discard semantic understanding, severely limiting practical autonomy. Alternative modular pipelines incorporate visual object detectors, but running open-vocabulary trackers on constrained onboard computers yields control frequencies too low for dynamic manipulation. To guarantee high-frequency state estimation, mainstream solutions often rely on high-performance off-board compute nodes or external motion capture systems~\cite{helpful_doggybot, humanoid_badminton}. While indispensable for extreme high-speed tasks (e.g., robotic sports), such infrastructure-heavy setups restrict the autonomy of legged systems in everyday spaces. Furthermore, while applying filtering algorithms is a standard remedy for low-frequency signals, utilizing them directly on a walking quadruped is notoriously difficult; traditional filters struggle to decouple the target's relative movement from the severe ego-motion of the floating base, leading to rapid drift. Consequently, many recent frameworks bypass the continuous floating-base control challenge by stopping the base first and relying primarily on quasi-static robotic arm operations~\cite{quadwbg, vbc, wildlma}.

To address these interconnected challenges, we propose SigLoMa, a fully onboard quadrupedal loco-manipulation framework tailored for unstructured, open-world scenarios. SigLoMa is built upon three core pillars. First, we introduce Sigma Points, a lightweight, category-agnostic geometric representation that bridges upstream open-vocabulary semantic detectors and downstream continuous control, enabling efficient simulation training without dense visual rendering. Second, we design an ego-centric Kalman Filter (KF) that bridges the frequency gap between low-rate embodied perception outputs and the high-frequency demands of continuous floating-base control. By explicitly compensating for camera ego-motion, the KF separates target motion from base motion and produces robust, high-rate state estimates. Finally, we train a robust RL policy with an Active Sampling Curriculum (ASC) and Hint Poses to learn object-specific approach trajectories without heavy reward engineering. To improve reliability during the final grasping phase, SigLoMa incorporates a memory-augmented formulation that mitigates structural blind spots and sim-to-real open-loop drift.

Our core contributions are summarized as follows:
\begin{enumerate}
	\item \textbf{A Hardware-Efficient Loco-Manipulation System}: SigLoMa is a fully onboard vision-based framework that enables precise floating-base control without external motion capture or off-board computation;
	\item \textbf{An Ego-Centric Geometric Perception Pipeline}: a state estimation architecture built on Sigma Points and an ego-centric KF, which bridges embodied semantic tracking and fast control through explicit ego-motion compensation;
	\item \textbf{A Robust Long-Horizon Learning Curriculum}: an integrated RL strategy that uses an ASC and hint poses, together with a temporal memory network, to overcome sparse rewards and improve robustness under structural visual occlusions.
\end{enumerate}

%% file: sections/02_related_work.tex
\section{Related Work}
\label{sec:related_work}

\noindent\textbf{Perception-Aware Legged Control.} 
Integrating perception into reinforcement learning has driven breakthroughs in legged locomotion. 
To mitigate the RGB sim-to-real gap, early works leveraged elevation or depth maps via privileged two-stage distillation (e.g., DAgger~\cite{dagger})~\cite{loquercio2022learning, reconstruct, zhuang2023robot, cheng2023parkour, miki2022learning, gangapurwala2022rloc, duan2024learning, chen2025vmts, wang2025more, vital2023}, occasionally incorporating LiDAR for enhanced foothold planning~\cite{omni_perception, perceptive_him, omni_stair}. 

Transitioning to loco-manipulation, however, necessitates both geometric and semantic understanding. 
Many systems address this by training with privileged simulated object states~\cite{eth_badminton, playful_doggybot, marl_soccer, dribblebot, goalkeeper, hitter_pingpong} and deploying task-specific detectors. 
Most notably, robotic badminton~\cite{eth_badminton} utilizes an Extended Kalman Filter (EKF) for shuttlecock trajectory estimation, but heavily relies on Multi-Sensor Fusion and CompSLAM to maintain a rigidly stable global coordinate system. 
While effective for structured sports, this strict reliance on drift-free global estimation increases deployment complexity and restricts open-world mobility. 
Furthermore, such stringent tracking precludes generic, open-vocabulary vision. 
Consequently, achieving long-horizon floating-base manipulation using exclusively low-frequency semantic detectors on constrained onboard compute remains unresolved. 

\noindent\textbf{Quadruped Loco-Manipulation Pipelines.}
Current loco-manipulation architectures generally fall into implicit learning frameworks and modular decoupled designs. 
Implicit approaches, such as end-to-end RL~\cite{vbc} and behavior-cloning via Action Chunking Transformers~\cite{wildlma}, directly map observations to actions. 
However, they suffer from sample inefficiency, heavy reliance on teleoperation, and their latent representations preclude integration with explicit high-frequency state estimators. 

Modular systems typically outfit the quadruped with a robotic arm~\cite{quadwbg, fu2023deep, roboduet, mlm, weept}, executing a ``Maps-then-reach'' strategy. 
While transferring the accuracy burden to a dedicated manipulator relaxes base tracking requirements, it introduces substantial hardware cost, payload constraints, and power overhead. 
A highly cost-effective alternative mounts a gripper directly onto the base, which mandates continuous, highly precise base regulation. 
Existing systems utilizing this morphology either restrict manipulation to task-specific dynamic interceptions (e.g., catching spheres)~\cite{playful_doggybot} or heavily rely on global external cameras and low-frequency perception~\cite{helpful_doggybot}. 

In contrast, SigLoMa maximizes this arm-free morphology using exclusively onboard, ego-centric perception. 
By providing the high-frequency, low-latency state estimation essential for stringent continuous base control, it circumvents both the hardware bloat of armed quadrupeds and the external infrastructure reliance of prior base-mounted pipelines. 

%% file: sections/03_method.tex
\section{Method}
\label{sec:method}

\input{figures/01_pipeline}

Designing an open-world quadrupedal pick-and-place solution presents dual overarching challenges. For simulation training, the RL policy must generalize across diverse object geometries to interface seamlessly with high-level Vision Language Models (VLMs). For real-world deployment, maintaining high manipulation precision on a highly dynamic floating base imposes strict latency and frequency requirements; even minor visual tracking delays translate into centimeter-level base deviation errors, frequently causing grasp failures. To address these issues, we propose SigLoMa, a hierarchical and highly adaptive framework: (1) prompt-driven VLM parsing for open-vocabulary target detection; (2) an ego-centric KF handling camera motion compensation, frequency amplification, and delay masking; and (3) a robust high-level RL policy utilizing computationally lightweight Sigma Points to formulate dynamic velocity and posture commands for reaching the optimal operational pose.

\subsection{Hierarchical System Overview}
As depicted in Fig.~\ref{fig:pipeline}, the SigLoMa pipeline initiates with a VLM processing raw user language commands to instantiate a \texttt{pick\_target} and a \texttt{place\_target}. We then feed these zero-shot anchors into an onboard open-vocabulary tracking model, which yields low-frequency semantic masks. Because raw visual masks are computationally prohibitive for large-scale simulation and highly susceptible to sim-to-real domain shifts, we distill this raw perception into a sparse set of geometric feature points, termed \textit{Sigma Points}. Operating on these sparse features, a Kalman Filter (KF) bridges the low-frequency visual updates to output high-frequency, low-latency target state estimates. Finally, these estimates are routed to the RL policy network to generate continuous velocity and posture commands, achieving precise grasping and placing maneuvers.

\subsection{Exteroceptive State Representation via Sigma Points}
To capture object geometry while maintaining computational efficiency, we represent target objects via Sigma Points---a sparse, mathematically derived set of representative features that effectively encapsulate bounding volume and spatial orientation.

Let $\mathcal{P} = \{{}^C\mathbf{p}_i \in \mathbb{R}^3\}_{i=1}^N$ be the dense object point cloud transformed into the camera coordinate frame $\{C\}$, with corresponding surface normals ${}^C\mathbf{n}_i$. We first extract the visible point set $\mathcal{V}$ based on strict geometric back-face culling and Field-of-View (FoV) constraints. The visibility condition dictates that the surface normal must oppose the viewing direction ${}^C\mathbf{p}_i$:
$$ \mathcal{V} = \left\{ {}^C\mathbf{p}_i \in \mathcal{P} \mid {}^C\mathbf{n}_i \cdot {}^C\mathbf{p}_i < 0 \land \text{FoV}({}^C\mathbf{p}_i) = \text{True} \right\} $$

In simulation, points uniformly cover 3D surfaces, whereas real-world points back-projected from 2D masks suffer from perspective-induced downsampling on slanted and distant faces. To bridge this sim-to-real density gap, we assign a geometry-aware weight $w_i$ to each visible point. This weight emulates the real camera pixel distribution by accounting for the differential solid angle subtended by surface patches~\cite{hartley2003multiple}: $w_i = \max \left(0, \frac{-{}^C\mathbf{n}_i \cdot {}^{C}\mathbf{p}_i}{\|{}^{C}\mathbf{p}_i\|^3} \right)$. By incorporating $w_i$ into a weighted Principal Component Analysis (PCA) on $\mathcal{V}$~\cite{pattern_recognition}, the extracted Sigma Points in simulation achieve a statistical distribution tightly aligned with real-world pixel-level observations. This alignment ensures the downstream RL policy remains invariant to the underlying point sampling source. The weighted centroid ${}^C\boldsymbol{\mu}$ and the spatial covariance matrix $\mathbf{\Sigma}_p$ are computed as:
$$ {}^C\boldsymbol{\mu} = \frac{\sum_{i \in \mathcal{V}} w_i {}^C\mathbf{p}_i}{\sum_{i \in \mathcal{V}} w_i}, \quad \mathbf{\Sigma}_p = \frac{\sum_{i \in \mathcal{V}} w_i ({}^C\mathbf{p}_i - {}^C\boldsymbol{\mu})({}^C\mathbf{p}_i - {}^C\boldsymbol{\mu})^T}{\sum_{i \in \mathcal{V}} w_i} $$
Through eigendecomposition $\mathbf{\Sigma}_p \mathbf{e}_k = \lambda_k \mathbf{e}_k$, we extract the eigenvalues $\lambda_k$ and eigenvectors $\mathbf{e}_k$ ($k \in \{1,2,3\}$). We systematically sample two points per principal axis alongside the geometric centroid, yielding a geometric feature set $\mathbf{S} = \{ \mathbf{s}_j \}_{j=0}^6$:
$$ {}^C\mathbf{s}_0 = {}^C\boldsymbol{\mu}, \quad {}^C\mathbf{s}_{k,\pm} = {}^C\boldsymbol{\mu} \pm \alpha \sqrt{\lambda_k} \mathbf{e}_k $$
where $\alpha$ is a scaling factor. To maintain temporal consistency with downstream filtering and control modules, we denote this observed point set at any given camera frame $t$ as ${}^{C_t}\mathbf{S}^{\text{obs}}_t$. During RL training, we analytically compute pseudo-visible surfaces relative to the camera frame. In the real world, semantic masks are back-projected into the 3D camera coordinate frame and undergo the identical PCA process to obtain the actual Sigma Points. The processing pipeline is illustrated in Fig.~\ref{fig:pipeline}.

\subsection{Ego-Centric State Estimation via Kalman Filtering}
Significant visual tracking latency severely restricts dynamic manipulation. Due to the large-range motion of the floating base, traditional methods often resort to external motion capture systems or only predict the trajectory of moving objects. To this end, we implement an ego-centric KF updating locally within the camera frame, utilizing Visual Odometry (VO) to mitigate base perturbation.

To track the target's geometric structure, we apply the filtering process uniformly to all elements of the Sigma Points set $\mathbf{S}$. For each Sigma Point $\mathbf{s}_{j}$ ($j \in \{0, \dots, 6\}$), the system state is defined directly in the current camera frame $\{C_t\}$ as ${}^{C_t}\mathbf{x}_{j,t} = [({}^{C_t}\mathbf{s}_{j,t})^T, ({}^{C_t}\mathbf{v}_{j,t})^T]^T \in \mathbb{R}^6$, where ${}^{C_t}\mathbf{s}_{j,t}$ and ${}^{C_t}\mathbf{v}_{j,t}$ represent its 3D position and relative velocity, respectively. The core innovation of our filter lies in decoupling the state transition into two distinct physical processes: point motion prediction and camera ego-motion compensation.

For the prediction step, we adopt a linear kinematic approximation to predict the $j$-th point's short-term displacement relative to the previous frame $\{C_{t-1}\}$:
\begin{equation}
    {}^{C_{t-1}}\mathbf{s}^{-}_{j,t} = {}^{C_{t-1}}\mathbf{s}_{j,t-1} + {}^{C_{t-1}}\mathbf{v}_{j,t-1} \Delta t
\end{equation}
Simultaneously, we rectify the shifts in the sensory frame caused by the robot's base movement. Given the sequential world-to-camera poses ${}^W\mathbf{T}_{C_{t-1}}$ and ${}^W\mathbf{T}_{C_t}$ provided by the VO, we extract the relative transformation ${}^{C_t}\mathbf{T}_{C_{t-1}} = ({}^W\mathbf{T}_{C_t})^{-1} {}^W\mathbf{T}_{C_{t-1}}$. Using the rotational component ${}^{C_t}\mathbf{R}_{C_{t-1}}$ and translational component ${}^{C_t}\mathbf{t}_{C_{t-1}}$ of this relative transformation, we map each predicted point state directly into the new camera perspective:
\begin{equation}
    {}^{C_t}\hat{\mathbf{s}}_{j,t} = {}^{C_t}\mathbf{R}_{C_{t-1}} {}^{C_{t-1}}\mathbf{s}^{-}_{j,t} + {}^{C_t}\mathbf{t}_{C_{t-1}}, \quad {}^{C_t}\hat{\mathbf{v}}_{j,t} = {}^{C_t}\mathbf{R}_{C_{t-1}} {}^{C_{t-1}}\mathbf{v}_{j,t-1}
\end{equation}
When the visual observation arrives, the newly computed Sigma Points from the PCA serve as measurement updates to correct their respective prediction errors. This filtering process yields the robust Sigma Points set ${}^{C_t}\mathbf{S}^{\text{obs}}_t$ required by the downstream RL policy. It abolishes the dependency on a tightly unified global coordinate system and bounds the residual error to the inter-frame increment, resulting in low-delay geometric state estimations. Detailed mathematical derivations of the entire filtering process are provided in Appendix \ref{sec:kalmanfilter}.

\input{figures/02_tasks}

\subsection{Policy Formulation and Robust Learning Framework}

\textbf{Task Taxonomy and MDP Setup.} We formulate the loco-manipulation problem across a discrete task space $\mathcal{T} = \{ \tau_{\text{long}}, \tau_{\text{short}}, \tau_{\text{release}} \}$, which is visually summarized in Fig.~\ref{fig:tasks}. Specifically, we deploy dozens of daily items and regular objects to evaluate the system. For picking operations, an object whose longest axis exceeds the maximum gripper aperture is classified as a long-axis target requiring $\tau_{\text{long}}$, which dictates an orthogonal approach trajectory; conversely, smaller targets default to $\tau_{\text{short}}$. For placing operations, the system universally executes the object release task $\tau_{\text{release}}$. 

To solve these tasks, our high-level RL policy maps observations to actions at each control step $t$. As illustrated in Fig.~\ref{fig:pipeline}, the observation space is decoupled into proprioception and exteroception. The proprioceptive state $\mathbf{o}^{\text{prop}}_t = [{}^B\mathbf{g}_t, {}^B\mathbf{v}_t, {}^B\boldsymbol{\omega}_t, \mathbf{a}_{t-1}, c_{\text{task}}]$ encapsulates the robot's base-projected gravity, linear and angular velocities, the previous action, and the task phase flag. The exteroceptive state relies entirely on the tracking of Sigma Points ${}^{C_t}\mathbf{S}^{\text{obs}}_t \in \mathbb{R}^{7 \times 3}$, which are processed via a dual-horizon architecture. Specifically, a short-horizon encoder processes a high-frequency history of $H_{\text{short}}$ frames to overcome perception noise, while a long-horizon encoder processes a low-frequency history of $H_{\text{long}}$ frames to provide a memory window of effective perception within blind spots. These extracted temporal features are fused with the proprioceptive state and passed to the Actor network, which outputs continuous base velocities and body pitch commands $\mathbf{a}_t = [v_x, v_y, \omega_{z}, \theta_{y}]^T$. Finally, a pre-trained low-level controller~\cite{slr} executes these commands. We train the policy using task constraints and stability regularizations to achieve optimal terminal alignment and stable locomotion, detailing formulations in Appendix \ref{sec:appendix-reward-function}.

\textbf{Trajectory Shaping and Active Sampling.} To combat the inherent sample inefficiency and sparse rewards of long-horizon tasks, we introduce two critical training optimizations. First, to regularize the policy's approach behaviors, we establish an optimal terminal pose for open-loop grasp triggering for each $\tau \in \mathcal{T}$. Since standard sparse rewards poorly constrain the strict approach trajectories dictated by object geometry (e.g., elongated objects mandate orthogonal approaches due to limited gripper aperture), we introduce \textit{hint poses} to implicitly regularize trajectory rollouts, as shown in Fig.~\ref{fig:tasks}. Serving as intermediate orientational waypoints, these hint poses enforce target pre-alignment, naturally shaping the approach to ensure smooth convergence to the optimal pose. 

Second, we utilize an Active Sampling Curriculum (ASC) that dynamically shifts the training focus from guided skill-bootstrapping to aggressive hard-negative mining. Inspired by biological learning, we design a \textit{near-optimal} initialization that functions as an active ``feeding'' mechanism. By intentionally spawning objects in close proximity to the optimal pose during early training, it provides immediate, dense rewards that drastically lower the initial exploration barrier, conceptually akin to parental nurturing. As the policy matures, this feeding mechanism naturally decays, seamlessly shifting the curriculum's focus toward a \textit{failure-replay} distribution to aggressively eliminate bottleneck behaviors. 

This curriculum transition is mathematically driven by a normalized task competency metric $\rho = \min(s / s_{\text{thresh}}, 1.0)$. Here, $s$ denotes the running success rate, evaluated via dual geometric boundaries: a rollout succeeds if the terminal pose deviation falls below the success boundary and is registered as a failure when it violates the failure boundary at timeout. The formal definitions of these boundaries are provided in Appendix~\ref{sec:task_criteria}. $s_{\text{thresh}}$ is a predefined curriculum threshold. The sampling probability $P_i$ for initialization type $i \in \{\text{near-optimal}, \text{failure-replay}\}$ transitions exponentially from $p_{i,\text{start}}$ to $p_{i,\text{end}}$:
$$P_i(\rho) = p_{i,\text{end}} + (p_{i,\text{start}} - p_{i,\text{end}}) \exp(-\lambda_{\text{ASC}} \rho)$$
where $\lambda_{\text{ASC}}$ governs the transition rate. Consequently, the feeding probability exponentially decays ($p_{i,\text{start}} > p_{i,\text{end}}$) while the failure-replay probability is exponentially prioritized ($p_{i,\text{start}} < p_{i,\text{end}}$).

\textbf{Memory-Augmented Blind-Spot Handling.} Terminal grasping inevitably pushes the target out of the camera's FoV. To robustly simulate the ensuing open-loop KF drift during this out-of-FoV phase, we inject a cumulative, clamped random-walk noise $\mathbf{d}_t \in \mathbb{R}^3$ uniformly into the true Sigma Points ${}^{C_t}\mathbf{S}^{\text{true}}_t$:
\begin{equation}
    \mathbf{d}_t = \text{clip}\big(\mathbf{d}_{t-1} + \mathcal{N}(\mathbf{0}, \sigma_{\text{drift}}^2 \mathbf{I}), -d_{\max}, d_{\max}\big), \quad {}^{C_t}\mathbf{S}^{\text{obs}}_t = {}^{C_t}\mathbf{S}^{\text{true}}_t + \mathbf{d}_t
\end{equation}
where $\sigma_{\text{drift}}$ is the standard deviation governing the random-walk variance, $d_{\max}$ restricts the maximum positional drift, and $\mathbf{d}_t$ resets to zero upon visual tracking recovery. To maintain high-precision manipulation through this blind spot, a sufficient temporal receptive field is essential. To circumvent the prohibitive overhead of buffering at the $50\mathrm{Hz}$ control rate, we maintain a lightweight $2\mathrm{s}$ perception memory down-sampled to $5\mathrm{Hz}$. Processed via a Temporal Convolutional Network (TCN), this historical buffer effectively encapsulates spatiotemporal context, natively bridging the sim-to-real blind-spot gap under strict onboard compute limits.

%% file: figures/01_pipeline.tex
\begin{figure*}[t]
    \centering
    \includegraphics[width=1.0\textwidth]{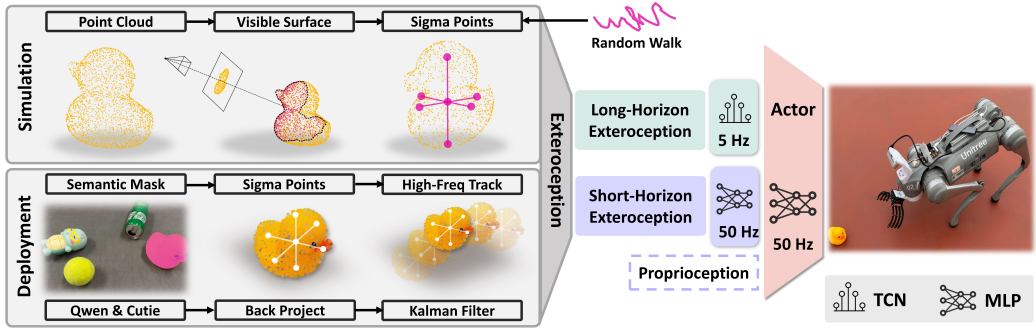}
    \caption{\textbf{Overall System Architecture.} \textbf{(1) Perception Design:} In simulation, 50~Hz Sigma Points are extracted from object point clouds via visible surface computation and PCA, with random-walk noise injected in blind spots to ensure robust real-world open-loop predictions. In deployment, semantic masks from visual tracking are back-projected to the camera frame to generate 5~Hz Sigma Points, which are then upsampled and latency-compensated to 50~Hz by an ego-centric Kalman Filter. \textbf{(2) Network Architecture:} Dual encoders process the Sigma Points across different frequencies and time horizons. Finally, an Actor network translates these temporal features into continuous locomotion and posture commands.}
    \label{fig:pipeline}
\end{figure*}

%% file: figures/02_tasks.tex
\begin{figure*}[t]
    \centering
    \includegraphics[width=1.0\textwidth]{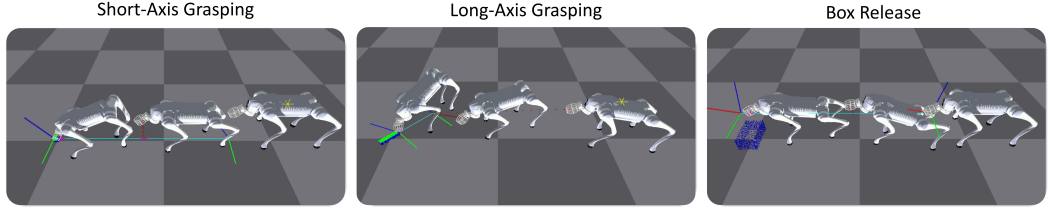}
    \caption{\textbf{Task Taxonomy and Pose Design.} Optimal terminal poses are located on the objects, while hint poses are positioned mid-way. The green lines connecting them indicate the suggested motion trajectories. Object point clouds are visualized as blue points, with real-time computed visible surfaces highlighted in green. Notably, for long-axis objects, the hint pose implicitly guides the robot to maneuver from the front to the side to approach the short-axis grasping point.}
    \label{fig:tasks}
\end{figure*}

%% file: sections/04_experiments.tex
\section{Experimental Results}
\label{sec:experiments}

\input{figures/03_real_exp}

\subsection{Simulation Setup and Ablation Studies}
The RL policy is trained using Proximal Policy Optimization (PPO)~\cite{schulman2017proximal} in Isaac Gym~\cite{lggym}. To ensure robust open-world generalization, the training environment incorporates a diverse object distribution, encompassing geometric primitives (e.g., cuboids, spheres, boxes, and cylinder baskets) and 27 everyday household items from the YCB dataset~\cite{calli2015ycb}. Furthermore, the initial 6-DoF poses of both the robot and the target objects are uniformly randomized at the onset of each episode. Despite this extensive morphological and spatial randomization, SigLoMa's compact Sigma Point representation and the ASC enable exceptional sample efficiency; the policy fully converges in approximately 6 hours on a single NVIDIA RTX 4090 GPU.

We evaluate the impact of SigLoMa's core algorithmic components through extensive ablation studies. Specifically, we investigate variants operating without hint poses (\textbf{No Hint}), without the ASC (\textbf{No ASC}), and with modified blind-spot memory architectures (e.g., completely removing the TCN encoder (\textbf{No TCN}), or substituting the TCN with an MLP (\textbf{Enc-MLP}) or a GRU (\textbf{Enc-GRU})). These configurations were evaluated across a diverse set of objects, mapped to task space $\mathcal{T}$.

\input{tables/01_sim_results}

Results indicate that the introduction of hint poses and the ASC serves as the primary catalyst for learning long-horizon, long-axis grasping maneuvers. Without these mechanisms, the policy consistently fails to explore successful perpendicular approach trajectories. Furthermore, ablating the temporal memory window or substituting the sequential encoder with an MLP demonstrably degrades grasping success. This indicates that retaining representations of previously valid visual observations is crucial for enabling the robot to maintain a stable posture. 

Moreover, the GRU variant underperforms the TCN. We attribute this to the GRU's recursive memory, which is highly sensitive to the non-stationary data distributions in RL. In contrast, the TCNs's feedforward structure, operating over a fixed receptive field without accumulating hidden state errors, inherently provides a more robust temporal representation. Interestingly, during testing, we observed emergent active-perception behaviors: the agent inherently maintains a ``head-up'' pitch at a distance to maximize FoV retention, performs granular lateral corrections upon proximity, and executes a smooth deceleration into the blind spot to minimize internal KF drift.

\subsection{Real-World Hardware Deployment}
We deployed SigLoMa on a Unitree Go2 quadruped, equipped with a RealSense D435i camera, a custom low-cost gripper, and an onboard NVIDIA Jetson Orin NX. VO is provided by Isaac ROS Visual SLAM at 60 Hz. For the perception stack, we leverage the Qwen 3.5 VLM API \cite{qwen35blog} to extract initial semantic masks via zero-shot reasoning. These masks initialize Cutie \cite{cuite}, which runs entirely onboard to provide continuous target tracking at a low frequency of ~5 Hz. Detailed hardware configurations and task execution workflows are deferred to the Appendix \ref{sec:hardware_workflow}.

As illustrated in Fig.~\ref{fig:real_exp}, we documented performance across three distinct pick-and-place tasks representing the SigLoMa taxonomy $\mathcal{T}$: \textit{ball-to-bin} ($\tau_{\text{short}} \to \tau_{\text{release}}$), \textit{toy-to-hamper} ($\tau_{\text{short}} \to \tau_{\text{release}}$), and \textit{water bottle-to-box} ($\tau_{\text{long}} \to \tau_{\text{release}}$). Each setting underwent 10 trials, from which we recorded the first-attempt end-to-end success counts. To evaluate the effectiveness of our proposed methods for closing the sim-to-real gap, we conducted ablation studies isolating the KF (\textbf{No KF}) and the blind-spot random-walk noise (\textbf{No BS-Noise}), ultimately comparing the SigLoMa autonomous pipeline against human teleoperation (\textbf{Teleop}).

The real-world executions definitively highlight the indispensability of SigLoMa's ego-centric state estimation framework. The KF effectively compensates for processing latency and bridges the low-frequency vision backbone, ensuring real-time locomotion reactivity. Furthermore, injecting random-walk noise during simulation training proves crucial for modeling the KF's open-loop prediction drift. This targeted augmentation explicitly reduces the sim-to-real gap, enabling robust policy execution even within visual blind spots. Ultimately, the fully onboard SigLoMa autonomous pipeline achieves highly precise floating-base maneuvers, yielding end-to-end success counts that perform on par with human teleoperation.

%% file: figures/03_real_exp.tex
\begin{figure*}[t]
    \centering
    \includegraphics[width=1.0\textwidth]{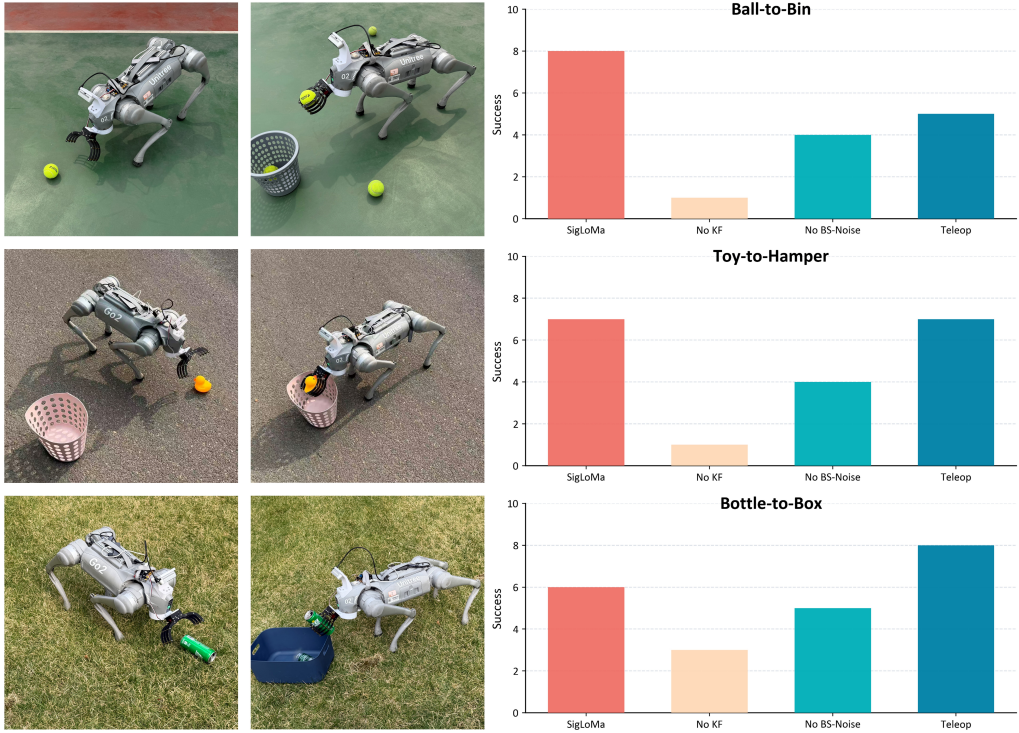}
    \caption{\textbf{Real-World Experimental Results.} Snapshots of the pick and place phases across three tasks, alongside their end-to-end success counts.}
    \label{fig:real_exp}
\end{figure*}

%% file: tables/01_sim_results.tex
\begin{table*}[htbp]
\centering
\scriptsize
\setlength{\tabcolsep}{4pt}
\renewcommand{\arraystretch}{1.3}
\caption{\footnotesize
Simulation Ablation Studies. Success is defined as the robot successfully stopping at the task-specific optimal pose (i.e., multi-dimensional tracking error strictly less than predefined success thresholds). We report the mean success rate ($\%$) and standard deviation for each condition. Each experiment is evaluated over 100 independent trials, and the reported values are averaged across 3 independent evaluation runs with randomized initial poses for both the robot and the object.
}
\label{tab:sim_results}
\begin{tabular}{l ccc@{\hskip 15pt}ccc@{\hskip 15pt}cc}

\toprule
& \multicolumn{3}{c}{\textbf{Short-Axis Grasp}} 
& \multicolumn{3}{c}{\textbf{Long-Axis Grasp}} 
& \multicolumn{2}{c}{\textbf{Target Release}} \\ 
\cmidrule(lr){2-4} \cmidrule(lr){5-7} \cmidrule(l){8-9}
\textbf{Method} & Cuboid & Apple & Sugar 
& Sphere & Conditioner & Shampoo 
& Basket & Box \\

\midrule
\textbf{SigLoMa} & $96.3_{\pm 1.5}$ & $\mathbf{97.0_{\pm 1.0}}$ & $\mathbf{96.0_{\pm 1.0}}$ & $\mathbf{87.7_{\pm 3.2}}$ & $83.3_{\pm 2.5}$ & $\mathbf{85.0_{\pm 2.0}}$ & $\mathbf{95.3_{\pm 1.5}}$ & $\mathbf{92.7_{\pm 2.1}}$ \\

\midrule
\textbf{Enc-MLP} & $\mathbf{96.7_{\pm 1.5}}$ & $96.0_{\pm 2.0}$ & $95.3_{\pm 2.1}$ & $83.3_{\pm 2.1}$ & $79.3_{\pm 2.5}$ & $82.3_{\pm 3.5}$ & $91.0_{\pm 1.7}$ & $87.3_{\pm 2.5}$ \\

\midrule
\textbf{Enc-GRU} & $82.7_{\pm 3.2}$ & $81.3_{\pm 3.5}$ & $83.3_{\pm 2.1}$ & $72.0_{\pm 3.6}$ & $67.7_{\pm 4.2}$ & $69.3_{\pm 3.1}$ & $82.3_{\pm 3.1}$ & $78.3_{\pm 3.5}$ \\

\midrule
\textbf{No TCN} & $89.3_{\pm 2.1}$ & $88.0_{\pm 2.6}$ & $90.0_{\pm 2.0}$ & $82.0_{\pm 2.6}$ & $\mathbf{86.3_{\pm 2.1}}$ & $83.3_{\pm 2.5}$ & $88.3_{\pm 2.1}$ & $84.3_{\pm 2.1}$ \\

\midrule
\textbf{No ASC} & $78.0_{\pm 2.0}$ & $76.3_{\pm 2.5}$ & $77.7_{\pm 2.1}$ & $44.0_{\pm 4.0}$ & $39.3_{\pm 4.5}$ & $43.3_{\pm 4.5}$ & $77.0_{\pm 3.0}$ & $73.0_{\pm 3.6}$ \\

\midrule
\textbf{No Hint} & $72.7_{\pm 2.5}$ & $71.3_{\pm 3.1}$ & $73.0_{\pm 2.6}$ & $16.0_{\pm 3.6}$ & $12.0_{\pm 2.0}$ & $14.3_{\pm 2.5}$ & $72.3_{\pm 2.5}$ & $68.3_{\pm 2.5}$ \\

\bottomrule
\end{tabular}
\end{table*}

%% file: sections/05_conclusion.tex
\section{Conclusion}
\label{sec:conclusion}
In this paper, we present SigLoMa, a low-cost, open-world solution for quadrupedal loco-manipulation that eliminates the need for external tracking and high-compute platforms. To achieve this, we introduce the Sigma Point representation for efficient RL and sim-to-real transfer, alongside an ego-centric Kalman Filter that bridges the frequency gap between semantic detection and continuous control. On the learning front, an ASC guided by hint poses and a TCN with random-walk noise accelerate multi-task convergence and robustly handle visual blind spots. Real-world experiments demonstrate that SigLoMa autonomously adapts grasping strategies to diverse objects, performing dynamic pick-and-place with success rates comparable to expert human teleoperation.

\subsection{Limitations and Future Work}
Despite strong sim-to-real transfer, SigLoMa exhibits a few notable limitations. First, manipulation is currently constrained to flat surfaces; future work will focus on integrating local elevation maps and training the policy from scratch to maintain the stability of the gripper's pose on uneven terrains. Second, acting primarily as a local visual servoing policy, the framework lacks long-horizon navigation capabilities, which could be addressed by incorporating a semantic-based navigation module to enable long-distance traversal and adaptive obstacle avoidance. Third, as a cascaded multi-module system, the overall grasping precision is susceptible to perception instability. Specifically, errors from frequency fluctuations or dropped frames in any single module can propagate downstream. To mitigate this, future efforts could explore optimizing system-level integration and enhancing the temporal memory architecture to build resilience against such sensory data fluctuations.

%% file: sections/99_references.tex
\bibliography{bib/references}

%% file: appendices/00_hardware_workflow.tex
\section{Hardware Setup and Task Workflow}
\label{sec:hardware_workflow}

\textbf{Hardware Setup.} Our hardware setup utilizes the open-source 3D-printed mounting brackets from \cite{helpful_doggybot} to firmly secure the camera and the end-effector. Specifically, as shown in Figure \ref{fig:setup}, an overhead D435i camera is mounted with a fixed pitch angle. The end-effector itself is an ultra-low-cost ($ < \$20$) servo-driven two-finger gripper.

\input{figures/00_setup}

\textbf{Task Workflow.} As illustrated in Figure~\ref{fig:workflow}, during the pick-and-place evaluation, the robot is initially teleoperated to scan the environment and record the approximate world coordinates of all target objects. The autonomous system then operates in a loop: it utilizes these global coordinates to orient until the specific target enters the visual field, which subsequently triggers the visual-based closed-loop policy. We deliberately omit complex navigation routing, as it falls outside the core scope of evaluating the visuo-motor manipulation policy.

\input{figures/05_workflow}

%% file: figures/00_setup.tex
\begin{figure}[htbp]
    \centering
    \includegraphics[width=0.48\textwidth]{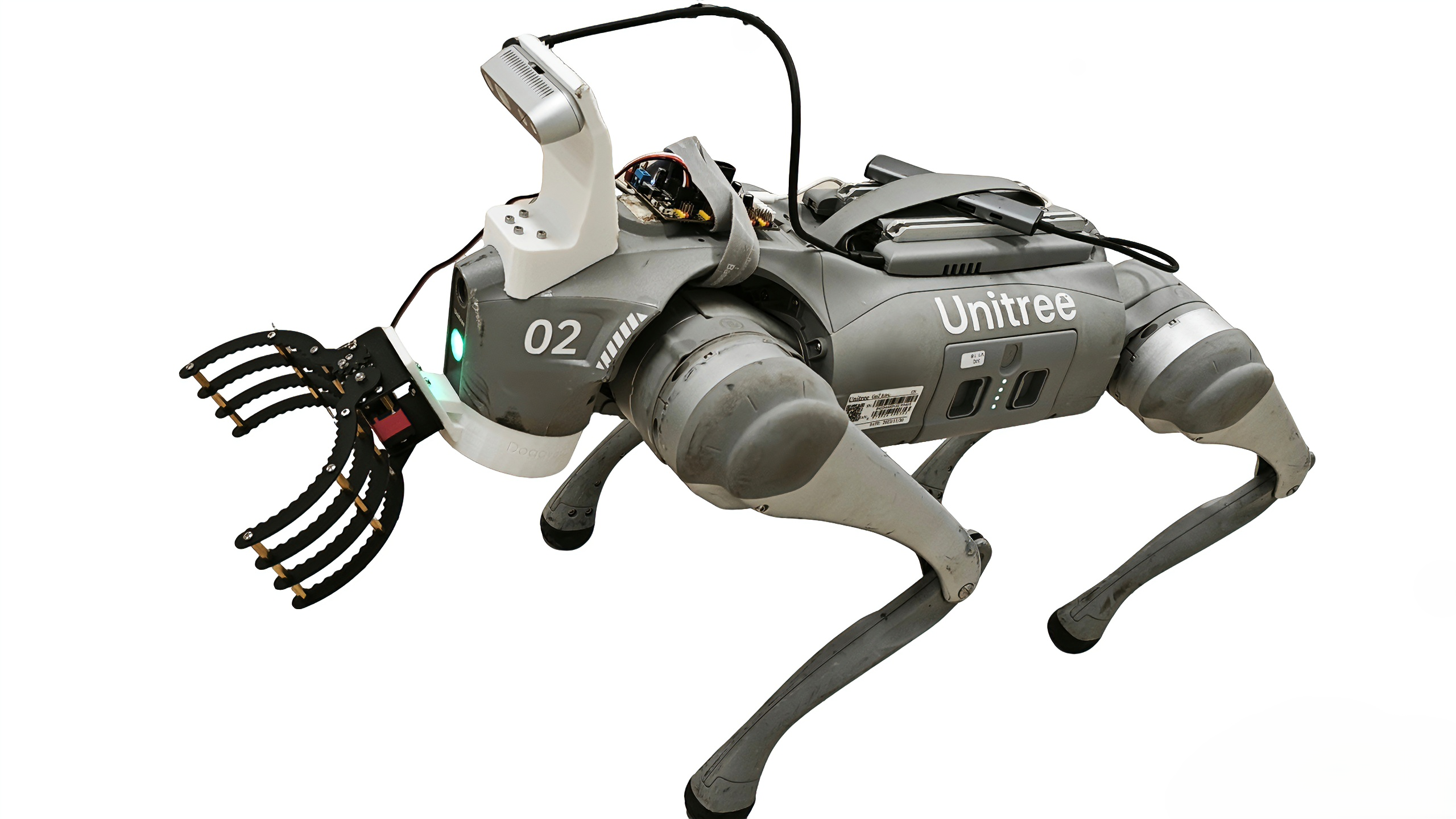}
    \caption{\textbf{Hardware Setup.} The overhead D435i camera is mounted at a fixed pitch alongside an ultra-low-cost servo-driven two-finger gripper using brackets \cite{helpful_doggybot}.}
    \label{fig:setup}
\end{figure}

%% file: figures/05_workflow.tex
\begin{figure}[htbp]
    \centering
    \includegraphics[width=0.9\linewidth]{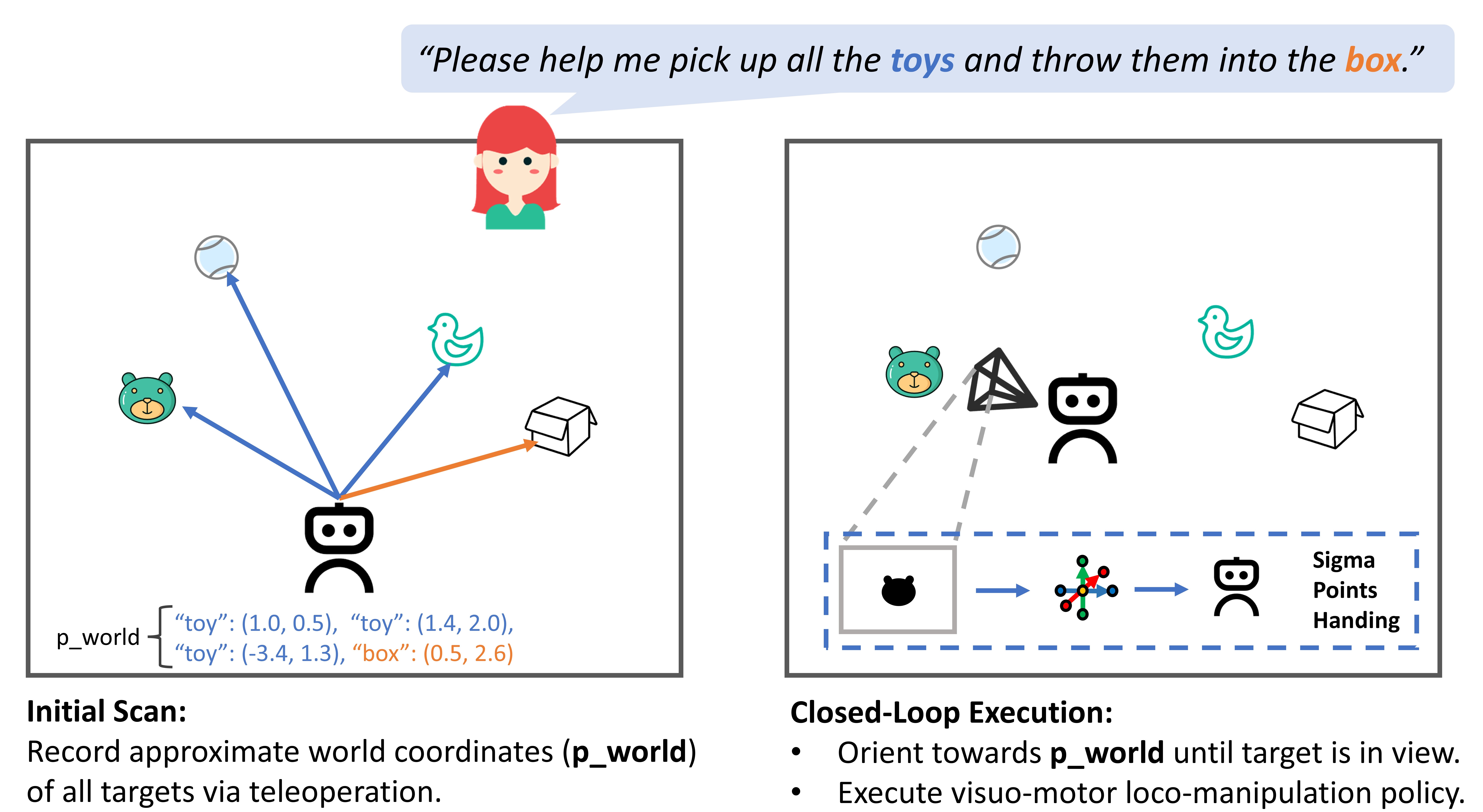}
    \caption{Overview of the task workflow. The robot first scans for approximate global coordinates, which are then used to orient the robot until the target is within the visual field to trigger the autonomous policy.}
    \label{fig:workflow}
\end{figure}

%% file: appendices/01_kalmanfilter.tex
\section{Comprehensive Formulation of the Ego-Centric KF}
\label{sec:kalmanfilter}
To provide a complete and systematic mathematical overview, this section details the state definition, process model, and measurement update of our ego-centric Kalman Filter.

\textbf{1. State Representation:} \\
The filter tracks the spatial state of each Sigma Point $\mathbf{s}_{j}$ ($j \in \{0, \dots, 6\}$) independently. The system state is defined directly in the current camera frame $\{C_t\}$ as a 6D vector encompassing its 3D position and relative velocity:
\begin{equation}
    {}^{C_t}\mathbf{x}_{j,t} = \begin{bmatrix} {}^{C_t}\mathbf{s}_{j,t} \\ {}^{C_t}\mathbf{v}_{j,t} \end{bmatrix} \in \mathbb{R}^6
\end{equation}

\textbf{2. Process Model and Ego-Motion Compensation:} \\
The state transition is decoupled into point motion prediction and camera ego-motion compensation. First, a linear kinematic model predicts the point's displacement relative to the previous frame $\{C_{t-1}\}$ over a time step $\Delta t$, where $\Delta t$ denotes the inter-frame interval:
\begin{equation}
    {}^{C_{t-1}}\mathbf{s}^{-}_{j,t} = {}^{C_{t-1}}\mathbf{s}_{j,t-1} + {}^{C_{t-1}}\mathbf{v}_{j,t-1} \Delta t
\end{equation}
Subsequently, we rectify the shifts in the sensory frame caused by the robot's base movement. Using the relative transformation ${}^{C_t}\mathbf{T}_{C_{t-1}} \in SE(3)$ provided by the Visual Odometry (VO), we extract its rotational component ${}^{C_t}\mathbf{R}_{C_{t-1}} \in SO(3)$ and translational component ${}^{C_t}\mathbf{t}_{C_{t-1}} \in \mathbb{R}^3$. The \textit{a priori} state estimate $\hat{\mathbf{x}}^{-}_{j,t}$ mapped to the new camera perspective $\{C_t\}$ is formulated as:
\begin{equation}
    {}^{C_t}\hat{\mathbf{s}}_{j,t} = {}^{C_t}\mathbf{R}_{C_{t-1}} {}^{C_{t-1}}\mathbf{s}^{-}_{j,t} + {}^{C_t}\mathbf{t}_{C_{t-1}}, \quad {}^{C_t}\hat{\mathbf{v}}_{j,t} = {}^{C_t}\mathbf{R}_{C_{t-1}} {}^{C_{t-1}}\mathbf{v}_{j,t-1}
\end{equation}

\textbf{3. Measurement Model and Dynamic Update:} \\
When a visual observation is available, we extract the empirical 3D position by back-projecting the segmented image pixels, denoting this spatial observation as $\mathbf{z}_{j,t} \in \mathbb{R}^3$. Because the measurement space explicitly isolates the positional subspace of the full state vector, the observation model is strictly linear:
\begin{equation}
    \mathbf{z}_{j,t} = \mathbf{H} {}^{C_t}\mathbf{x}_{j,t} + \boldsymbol{\nu}_t, \quad \text{with} \quad \mathbf{H} = [\mathbf{I}_{3 \times 3} \quad \mathbf{0}_{3 \times 3}]
\end{equation}
where $\boldsymbol{\nu}_t \sim \mathcal{N}(\mathbf{0}, \mathbf{R}_t)$ represents the measurement noise. 

To handle depth-dependent projection errors in physical deployments, the measurement covariance matrix $\mathbf{R}_t$ is dynamically scaled relative to the estimated $Z$-axis depth (i.e., the distance along the optical axis from the camera to the observed point) of the observed point:
\begin{equation}
    \mathbf{R}_t = \text{diag}(\sigma_X^2, \sigma_Y^2, \sigma_Z^2), \quad \text{where} \quad \sigma_X = \frac{Z}{f_x} \sigma_u, \quad \sigma_Y = \frac{Z}{f_y} \sigma_v, \quad \sigma_Z = \sigma_z
\end{equation}
with $f_x$ and $f_y$ denoting the intrinsic camera focal lengths (in pixels), $\sigma_u$ and $\sigma_v$ representing the pixel measurement uncertainties (in pixels), and $\sigma_z$ being a constant depth uncertainty (in meters). This dynamic scaling explicitly degrades the filter's trust in distant, noisy observations.

Finally, the posterior state estimate is optimally corrected via the Kalman gain $\mathbf{K}_t$:
\begin{equation}
    \mathbf{K}_t = \mathbf{P}^{-}_t \mathbf{H}^T (\mathbf{H} \mathbf{P}^{-}_t \mathbf{H}^T + \mathbf{R}_t)^{-1}
\end{equation}
\begin{equation}
    \hat{\mathbf{x}}^{+}_{j,t} = \hat{\mathbf{x}}^{-}_{j,t} + \mathbf{K}_t (\mathbf{z}_{j,t} - \mathbf{H} \hat{\mathbf{x}}^{-}_{j,t})
\end{equation}
where $\mathbf{P}^{-}_t$ is the prior error covariance matrix updated during the prediction step.

%% file: appendices/02_task_criteria.tex
\section{Geometric Alignment and Task Evaluation Criteria}
\label{sec:task_criteria}

This section defines the geometric alignment errors and terminal-state criteria used by both reward gating and ASC evaluation.

\textbf{Alignment Error Formulation.} Let $\mathbf{p}_e \in \mathbb{R}^3$ and $\mathbf{\Theta}_e \in (-\pi, \pi]^3$ denote the end-effector position and Euler angles (roll, pitch, yaw), and let $\mathbf{p}_{\text{opt}} \in \mathbb{R}^3$ and $\mathbf{\Theta}_{\text{opt}} \in (-\pi, \pi]^3$ denote the target optimal pose. The alignment errors are defined by weighted squared norms:
\begin{equation}
    e_{\text{pos}}^2 = (\mathbf{p}_e - \mathbf{p}_{\text{opt}})^\top \mathbf{W}_p (\mathbf{p}_e - \mathbf{p}_{\text{opt}}), \quad 
    e_{\text{rot}}^2 = (\mathbf{\Theta}_e - \mathbf{\Theta}_{\text{opt}})^\top \mathbf{W}_r (\mathbf{\Theta}_e - \mathbf{\Theta}_{\text{opt}})
\end{equation}
where $\mathbf{W}_p, \mathbf{W}_r \in \mathbb{R}^{3 \times 3}$ are diagonal matrices specifying per-axis relative weights.

Let $\mathbf{p}_{\text{hint}} \in \mathbb{R}^3$ be the intermediate hint pose. The cross-track error to the hint-to-optimal segment is
\begin{equation}
    d_{\text{path}} = \| \mathbf{p}_e - \text{Proj}(\mathbf{p}_e, \overline{\mathbf{p}_{\text{hint}}\mathbf{p}_{\text{opt}}}) \|_2
\end{equation}
where $\text{Proj}(\mathbf{p}_e, \overline{\mathbf{p}_{\text{hint}}\mathbf{p}_{\text{opt}}})$ denotes the orthogonal projection of $\mathbf{p}_e$ onto the line segment connecting $\mathbf{p}_{\text{hint}}$ and $\mathbf{p}_{\text{opt}}$.

\textbf{Terminal State Boundaries.} Define $\Delta \mathbf{p} = \mathbf{p}_e - \mathbf{p}_{\text{opt}}$ and $\Delta \mathbf{\Theta} = \mathbf{\Theta}_e - \mathbf{\Theta}_{\text{opt}}$. Success and failure indicators are written as
\begin{equation}
\mathds{1}_{\text{success}} = \mathds{1}\!\left( |\Delta p_x| < \epsilon_{x},\ |\Delta p_y| < \epsilon_{y},\ |\Delta \Theta_{\text{yaw}}| < \epsilon_{\text{yaw}},\ |\Delta \Theta_{\text{pitch}}| < \epsilon_{\text{pitch}} \right),
\end{equation}
\begin{equation}
\mathds{1}_{\text{fail}} = \mathds{1}\!\left( \text{timeout} \land \left[ |\Delta p_x| \ge \delta_{x} \lor |\Delta p_y| \ge \delta_{y} \lor |\Delta \Theta_{\text{yaw}}| \ge \delta_{\text{yaw}} \lor |\Delta \Theta_{\text{pitch}}| \ge \delta_{\text{pitch}} \right] \right).
\end{equation}
These indicators are used to compute the ASC running success rate $s$ and to gate terminal reward activation.

\begin{itemize}
    \item \textit{Success State ($\mathds{1}_{\text{success}}$):} Activated when the end-effector simultaneously satisfies translational and rotational tolerances relative to the optimal pose.
    
    \item \textit{Failure State ($\mathds{1}_{\text{fail}}$):} Activated at timeout when at least one positional or rotational error exceeds its failure bound.
\end{itemize}

All threshold constants and related hyperparameters are listed in Appendix~\ref{sec:appendix-train-config}.

%% file: appendices/03_reward_function.tex
\begingroup
\setlength{\intextsep}{5pt}
\setlength{\textfloatsep}{6pt}
\setlength{\abovecaptionskip}{3pt}
\setlength{\belowcaptionskip}{0pt}

\section{Reward Function}
\label{sec:appendix-reward-function}
To avoid complex heuristic shaping, the reward combines geometric path-following, sparse indicator terms, and stability penalties. Task-level terms guide hint-based tracking, terminal alignment, and field-of-view consistency, while regularization terms suppress wobbling and aggressive actions. We use a unified shaping term $E(\cdot)=\exp(-\|\cdot\|^2/\sigma_{\text{track}})$. Detailed geometric definitions are provided in Appendix~\ref{sec:task_criteria}.
\input{tables/03_reward_function}

%% file: tables/03_reward_function.tex
\begin{table}[H]
  \centering
  \small
  \setlength{\tabcolsep}{5pt}
  \renewcommand{\arraystretch}{0.92}
  \caption{Reward function components used in training.}
  \label{table:reward_functions}
  \begin{tabular}{lllc}
    \toprule
    \textbf{Category} & \textbf{Term} & \textbf{Expression} & \textbf{Weight} \\
    \midrule
    
    \multirow[c]{3}{*}{\begin{tabular}{@{}l@{}}\textit{Task \& Auxiliary} \\ \textit{Constraints}\end{tabular}} 
    & $r_{\text{hint}}$ & $E(d_{\text{path}}) \cdot E(e_{\text{rot}}) \cdot (1 + k E(e_{\text{pos}}))$ & 0.4 \\
    \noalign{\vskip 0.3mm}
    & $r_{\text{opt}}$ & $\mathds{1}_{\text{success}} \cdot E(e_{\text{pos}}) \cdot E(e_{\text{rot}}) \cdot E(\mathbf{v}_{\text{base}})$ & 20.0 \\
    \noalign{\vskip 0.3mm}
    & $r_{\text{miss}}$ & $\mathds{1}_{\text{out\_FOV}}$ & -0.1 \\
    \midrule
    
    \multirow{4}{*}{\begin{tabular}{@{}l@{}}\textit{Stability \& Control} \\ \textit{Regularizations}\end{tabular}} 
    & $r_{\text{roll}}$ & $g_y^2$ & -2.0 \\
    \noalign{\vskip 0.2mm}
    & $r_{\text{ang}}$ & $\|\boldsymbol{\omega}_{xy}\|^2$ & -0.1 \\
    \noalign{\vskip 0.2mm}
    & $r_{\text{smooth}}$ & $\|\mathbf{a}_t - \mathbf{a}_{t-1}\|^2$ & -0.01 \\
    \noalign{\vskip 0.2mm}
    & $r_{\text{limit}}$ & $\|\mathbf{a}_{\text{clip}} - \mathbf{a}\|^2$ & -0.1 \\
    \bottomrule
  \end{tabular}
\end{table}

%% file: appendices/04_kalman_config.tex
\section{Kalman Filter Configuration}
\label{sec:appendix-kalman-config}
\input{tables/04_kalman_config}

%% file: tables/04_kalman_config.tex
\begin{table}[H]
\centering
\small
\setlength{\tabcolsep}{5pt}
\renewcommand{\arraystretch}{0.92}
\caption{Ego-centric Kalman filter parameters.}
\label{tab:kf_params}
\begin{tabular}{llc}
\toprule
\textbf{Category} & \textbf{Parameter} & \textbf{Value} \\
\midrule
\textit{Process Noise ($\mathbf{Q}$)} & Position variance & $1.0 \times 10^{-6}$ $\text{m}^2$ \\
& Velocity variance & $1.0 \times 10^{-5}$ $\text{m}^2/\text{s}^2$ \\
\midrule
\textit{Measurement Noise ($\mathbf{R}$)} & Pixel std. dev. ($\sigma_u$, $\sigma_v$) & 20.0 px \\
& Depth base std. dev. ($\sigma_z$) & 0.05 m \\
\midrule
\textit{State Initialization ($\mathbf{P}_0$)} & Initial pos. variance & $1.0 \times 10^{-2}$ $\text{m}^2$ \\
& Initial vel. variance & $1.0 \times 10^{-1}$ $\text{m}^2/\text{s}^2$ \\
\bottomrule
\end{tabular}
\end{table}

%% file: appendices/05_domain_randomization.tex
\section{Domain Randomization}
\label{sec:appendix-domain-random}
\input{tables/04_domain_randomization}
\endgroup

%% file: tables/04_domain_randomization.tex
\begin{table}[H]
\centering
\small
\setlength{\tabcolsep}{5pt}
\renewcommand{\arraystretch}{0.92}
\caption{Domain randomization and observation noise parameters. $\mathcal{U}(a,b)$ denotes uniform distribution and $\mathcal{N}(0,\sigma)$ denotes zero-mean Gaussian noise.}
\label{tab:domain_rand}
\begin{tabular*}{\linewidth}{@{\extracolsep{\fill}} l l c @{}}
\toprule
\textbf{Category} & \textbf{Parameter} & \textbf{Distribution / Range} \\
\midrule
\multirow{4}{*}{\textit{Dynamics \& Geometry}} 
& Ground friction & $\mathcal{U}(0.2, 5.0)$ \\
& Restitution & $\mathcal{U}(0.0, 1.0)$ \\
& Base added mass & $\mathcal{U}(-1.0, 2.0)$ kg \\
& Sigma Points physical scale ($\alpha$) & $\mathcal{U}(1.0, 1.5)$ \\
\midrule
\multirow{4}{*}{\textit{Camera Extrinsics}} 
& Translation ($x, z$) & $\mathcal{U}(-0.02, 0.02)$ m \\
& Translation ($y$) & $\mathcal{U}(-0.005, 0.005)$ m \\
& Orientation (roll, yaw) & $\mathcal{U}(-0.5^\circ, 0.5^\circ)$ \\
& Orientation (pitch) & $\mathcal{U}(-2.0^\circ, 2.0^\circ)$ \\
\midrule
\multirow{6}{*}{\textit{Observation Noise}} 
& Perception delay & $\mathcal{U}(0, 50)$ ms \\
& Base linear velocity & $\mathcal{N}(0, 0.1)$ m/s \\
& Base angular velocity & $\mathcal{N}(0, 0.1)$ rad/s \\
& Projected gravity & $\mathcal{N}(0, 0.1)$ \\
& Sigma Points scale noise & $\mathcal{N}(0, 0.1)$ \\
& Sigma Points rotation noise & $\mathcal{N}(0, 0.1)$ rad \\
\bottomrule
\end{tabular*}
\end{table}

%% file: appendices/06_train_config.tex
\FloatBarrier
\clearpage
\section{Training Configuration}
\label{sec:appendix-train-config}
\input{tables/06_train_config}

%% file: tables/06_train_config.tex
\begin{table}[H]
\centering
\small
\renewcommand{\arraystretch}{0.95}
\caption{PPO Training and Network Configurations}
\label{tab:rl_config}
\begin{tabular}{llc}
\toprule
\textbf{Category} & \textbf{Parameter} & \textbf{Value} \\
\midrule
\multirow{8}{*}{\textit{RL Hyperparameters (PPO)}} 
& Number of environments & 4096 \\
& Batch size & 32768 \\
& Minibatch size & 8192 \\
& Learning rate & $3.0 \times 10^{-4}$ \\
& Discount factor ($\gamma$) & 0.99 \\
& GAE parameter ($\lambda$) & 0.95 \\
& PPO clip range & 0.2 \\
& Entropy coefficient & 0.01 \\
\midrule
\multirow{8}{*}{\textit{NN Architectures}} 
& Actor MLP [dims] & [512, 256, 128] \\
& Critic MLP [dims] & [512, 256, 128] \\
& TCN Encoder [channels] & [32, 32, 32] \\
& TCN kernel size & 5 \\
& TCN dropout & 0.1 \\
& Activation function & ELU \\
& Short-horizon frames ($H_{\text{short}}$) & 5 \\
& Long-horizon frames ($H_{\text{long}}$) & 10 \\
\midrule
\multirow{8}{*}{\shortstack[l]{\textit{Active Sampling} \\ \textit{\& Blind-Spot}}} 
& Curriculum threshold ($s_{\text{thresh}}$) & 0.15 \\
& Transition decay rate ($\lambda_{\text{ASC}}$) & 5.0 \\
& Initial feeding prob. ($p_{\text{near-optimal}, \text{start}}$) & 0.8 \\
& Final feeding prob. ($p_{\text{near-optimal}, \text{end}}$) & 0.1 \\
& Initial replay prob. ($p_{\text{failure-replay}, \text{start}}$) & 0.2 \\
& Final replay prob. ($p_{\text{failure-replay}, \text{end}}$) & 0.5 \\
& Drift standard deviation ($\sigma_{\text{drift}}$) [m] & 0.01 \\
& Max positional drift ($d_{\max}$) [m] & 0.10 \\
\midrule
\multirow{6}{*}{\textit{Terminal Criteria}} 
& Success bound: $\epsilon_x$, $\epsilon_y$ [m] & 0.05, 0.03 \\
& Success yaw bound: $\epsilon_{\text{yaw}}$ [rad] & 0.10 \\
& Success pitch bound: $\epsilon_{\text{pitch}}$ [rad] & 0.15 \\
& Failure bound: $\delta_x$, $\delta_y$ [m] & 0.10, 0.10 \\
& Failure yaw bound: $\delta_{\text{yaw}}$ [rad] & 0.20 \\
& Failure pitch bound: $\delta_{\text{pitch}}$ [rad] & 0.20 \\
\midrule
\multirow{4}{*}{\shortstack[l]{\textit{Rewards \& Actions}}} 
& Tracking sigma: $\sigma_{\text{track}}$ & 0.04 \\
& Hint-to-goal gain: $k$ & 1.0 \\
& Action clip ($v_x, v_y, \omega_z$) [m/s] & [-0.5, 0.5] \\
& Action clip (pitch) [rad] & [-$\pi$/6, $\pi$/6] \\
\bottomrule
\end{tabular}
\end{table}

%% file: bib/references.bib
@article{eth_badminton,
   title={Learning coordinated badminton skills for legged manipulators},
   volume={10},
   ISSN={2470-9476},
   url={http://dx.doi.org/10.1126/scirobotics.adu3922},
   DOI={10.1126/scirobotics.adu3922},
   number={102},
   journal={Science Robotics},
   publisher={American Association for the Advancement of Science (AAAS)},
   author={Ma, Yuntao and Cramariuc, Andrei and Farshidian, Farbod and Hutter, Marco},
   year={2025},
   month=may }

@misc{quadwbg,
      title={QuadWBG: Generalizable Quadrupedal Whole-Body Grasping}, 
      author={Jilong Wang and Javokhirbek Rajabov and Chaoyi Xu and Yiming Zheng and He Wang},
      year={2025},
      eprint={2411.06782},
      archivePrefix={arXiv},
      primaryClass={cs.RO},
      url={https://arxiv.org/abs/2411.06782}, 
}

@misc{roboduet,
      title={RoboDuet: Learning a Cooperative Policy for Whole-body Legged Loco-Manipulation}, 
      author={Guoping Pan and Qingwei Ben and Zhecheng Yuan and Guangqi Jiang and Yandong Ji and Shoujie Li and Jiangmiao Pang and Houde Liu and Huazhe Xu},
      year={2025},
      eprint={2403.17367},
      archivePrefix={arXiv},
      primaryClass={cs.RO},
      url={https://arxiv.org/abs/2403.17367}, 
}

@misc{vbc,
      title={Visual Whole-Body Control for Legged Loco-Manipulation}, 
      author={Minghuan Liu and Zixuan Chen and Xuxin Cheng and Yandong Ji and Ri-Zhao Qiu and Ruihan Yang and Xiaolong Wang},
      year={2024},
      eprint={2403.16967},
      archivePrefix={arXiv},
      primaryClass={cs.RO},
      url={https://arxiv.org/abs/2403.16967}, 
}

@misc{weept,
      title={Whole-body End-Effector Pose Tracking}, 
      author={Tifanny Portela and Andrei Cramariuc and Mayank Mittal and Marco Hutter},
      year={2025},
      eprint={2409.16048},
      archivePrefix={arXiv},
      primaryClass={cs.RO},
      url={https://arxiv.org/abs/2409.16048}, 
}

@misc{wildlma,
      title={WildLMa: Long Horizon Loco-Manipulation in the Wild}, 
      author={Ri-Zhao Qiu and Yuchen Song and Xuanbin Peng and Sai Aneesh Suryadevara and Ge Yang and Minghuan Liu and Mazeyu Ji and Chengzhe Jia and Ruihan Yang and Xueyan Zou and Xiaolong Wang},
      year={2025},
      eprint={2411.15131},
      archivePrefix={arXiv},
      primaryClass={cs.RO},
      url={https://arxiv.org/abs/2411.15131}, 
}

@misc{marl_soccer,
      title={Toward Real-World Cooperative and Competitive Soccer with Quadrupedal Robot Teams}, 
      author={Zhi Su and Yuman Gao and Emily Lukas and Yunfei Li and Jiaze Cai and Faris Tulbah and Fei Gao and Chao Yu and Zhongyu Li and Yi Wu and Koushil Sreenath},
      year={2025},
      eprint={2505.13834},
      archivePrefix={arXiv},
      primaryClass={cs.RO},
      url={https://arxiv.org/abs/2505.13834}, 
}

@misc{hitter_pingpong,
      title={HITTER: A HumanoId Table TEnnis Robot via Hierarchical Planning and Learning}, 
      author={Zhi Su and Bike Zhang and Nima Rahmanian and Yuman Gao and Qiayuan Liao and Caitlin Regan and Koushil Sreenath and S. Shankar Sastry},
      year={2025},
      eprint={2508.21043},
      archivePrefix={arXiv},
      primaryClass={cs.RO},
      url={https://arxiv.org/abs/2508.21043}, 
}

@misc{goalkeeper,
      title={Creating a Dynamic Quadrupedal Robotic Goalkeeper with Reinforcement Learning}, 
      author={Xiaoyu Huang and Zhongyu Li and Yanzhen Xiang and Yiming Ni and Yufeng Chi and Yunhao Li and Lizhi Yang and Xue Bin Peng and Koushil Sreenath},
      year={2022},
      eprint={2210.04435},
      archivePrefix={arXiv},
      primaryClass={cs.RO},
      url={https://arxiv.org/abs/2210.04435}, 
}

@misc{mlm,
      title={MLM: Learning Multi-task Loco-Manipulation Whole-Body Control for Quadruped Robot with Arm}, 
      author={Xin Liu and Bida Ma and Chenkun Qi and Yan Ding and Nuo Xu and Zhaxizhuoma and Guorong Zhang and Pengan Chen and Kehui Liu and Zhongjie Jia and Chuyue Guan and Yule Mo and Jiaqi Liu and Feng Gao and Jiangwei Zhong and Bin Zhao and Xuelong Li},
      year={2025},
      eprint={2508.10538},
      archivePrefix={arXiv},
      primaryClass={cs.RO},
      url={https://arxiv.org/abs/2508.10538}, 
}

@inproceedings{zhuang2023robot,
  author    = {Zhuang, Ziwen and Fu, Zipeng and Wang, Jianren and Atkeson, Christopher and Schwertfeger, S{\"o}ren and Finn, Chelsea and Zhao, Hang},
  title     = {Robot Parkour Learning},
  booktitle = {Conference on Robot Learning ({CoRL})},
  year      = {2023}
}

@article{cheng2023parkour,
  title   = {Extreme Parkour with Legged Robots},
  author  = {Cheng, Xuxin and Shi, Kexin and Agarwal, Ananye and Pathak, Deepak},
  journal = {arXiv preprint arXiv:2309.14341},
  year    = {2023}
}

@misc{reasan,
      title={REASAN: Learning Reactive Safe Navigation for Legged Robots}, 
      author={Qihao Yuan and Ziyu Cao and Ming Cao and Kailai Li},
      year={2025},
      eprint={2512.09537},
      archivePrefix={arXiv},
      primaryClass={cs.RO},
      url={https://arxiv.org/abs/2512.09537}, 
}

@misc{dribblebot,
      title={DribbleBot: Dynamic Legged Manipulation in the Wild}, 
      author={Yandong Ji and Gabriel B. Margolis and Pulkit Agrawal},
      year={2023},
      eprint={2304.01159},
      archivePrefix={arXiv},
      primaryClass={cs.RO},
      url={https://arxiv.org/abs/2304.01159}, 
}

@misc{helpful_doggybot,
      title={Helpful DoggyBot: Open-World Object Fetching using Legged Robots and Vision-Language Models}, 
      author={Qi Wu and Zipeng Fu and Xuxin Cheng and Xiaolong Wang and Chelsea Finn},
      year={2024},
      eprint={2410.00231},
      archivePrefix={arXiv},
      primaryClass={cs.RO},
      url={https://arxiv.org/abs/2410.00231}, 
}

@misc{dagger,
      title={A Reduction of Imitation Learning and Structured Prediction to No-Regret Online Learning}, 
      author={Stephane Ross and Geoffrey J. Gordon and J. Andrew Bagnell},
      year={2011},
      eprint={1011.0686},
      archivePrefix={arXiv},
      primaryClass={cs.LG},
      url={https://arxiv.org/abs/1011.0686}, 
}

@misc{playful_doggybot,
      title={Playful DoggyBot: Learning Agile and Precise Quadrupedal Locomotion}, 
      author={Xin Duan and Ziwen Zhuang and Hang Zhao and Soeren Schwertfeger},
      year={2025},
      eprint={2409.19920},
      archivePrefix={arXiv},
      primaryClass={cs.RO},
      url={https://arxiv.org/abs/2409.19920}, 
}

@article{loquercio2022learning,
  title={Learning visual locomotion with cross-modal supervision},
  author={Loquercio, Antonio and Kumar, Ashish and Malik, Jitendra},
  journal={arXiv preprint arXiv:2211.03785},
  year={2022}
}

@misc{reconstruct,
      title={Neural Scene Representation for Locomotion on Structured Terrain}, 
      author={David Hoeller and Nikita Rudin and Christopher Choy and Animashree Anandkumar and Marco Hutter},
      year={2022},
      eprint={2206.08077},
      archivePrefix={arXiv},
      primaryClass={cs.RO},
      url={https://arxiv.org/abs/2206.08077}, 
}

@article{miki2022learning,
  title={Learning robust perceptive locomotion for quadrupedal robots in the wild},
  author={Miki, Takahiro and Lee, Joonho and Hwangbo, Jemin and Wellhausen, Lorenz and Koltun, Vladlen and Hutter, Marco},
  journal={Science Robotics},
  volume={7},
  number={62},
  pages={eabk2822},
  year={2022},
  publisher={American Association for the Advancement of Science}
}

@inproceedings{yu2022visual,
  title={Visual-locomotion: Learning to walk on complex terrains with vision},
  author={Yu, Wenhao and Jain, Deepali and Escontrela, Alejandro and Iscen, Atil and Xu, Peng and Coumans, Erwin and Ha, Sehoon and Tan, Jie and Zhang, Tingnan},
  booktitle={Conference on Robot Learning},
  pages={1691--1702},
  year={2022},
  organization={PMLR}
}

@inproceedings{agarwal2023legged,
  title={Legged Locomotion in Challenging Terrains using Egocentric Vision},
  author={Agarwal, Ananye and Kumar, Ashish and Malik, Jitendra and Pathak, Deepak},
  booktitle={Conference on Robot Learning},
  pages={403--415},
  year={2023},
  organization={PMLR}
}

@article{gangapurwala2022rloc,
  title={RLOC: Terrain-aware legged locomotion using reinforcement learning and optimal control},
  author={Gangapurwala, Siddhant and Geisert, Mathieu and Orsolino, Romeo and Fallon, Maurice and Havoutis, Ioannis},
  journal={IEEE Transactions on Robotics},
  volume={38},
  number={5},
  pages={2908--2927},
  year={2022},
  publisher={IEEE}
}

@article{eth2025robust,
  title={Robust Reinforcement Learning-Based Locomotion for Resource-Constrained Quadrupeds with Exteroceptive Sensing},
  author={ETH-PBL},
  journal={arXiv preprint arXiv:2505.12537},
  year={2025}
}

@inproceedings{duan2024learning,
  title={Learning vision-based bipedal locomotion for challenging terrain},
  author={Duan, Helei and Pandit, Bikram and Gadde, Mohitvishnu S and Van Marum, Bart Jaap and Dao, Jeremy and Kim, Chanho and Fern, Alan},
  booktitle={2024 IEEE International Conference on Robotics and Automation (ICRA)},
  pages={56--62},
  year={2024},
  organization={IEEE}
}

@article{chen2025vmts,
  title={VMTS: Vision-Assisted Teacher-Student Reinforcement Learning for Multi-Terrain Locomotion in Bipedal Robots},
  author={Chen, Fu and Wan, Rui and Liu, Peidong and Zheng, Nanxing and Zhou, Bo},
  journal={arXiv preprint arXiv:2503.07049},
  year={2025}
}

@article{wang2025more,
  title={MoRE: Mixture of Residual Experts for Humanoid Lifelike Gaits Learning on Complex Terrains},
  author={Wang, Dewei and Wang, Xinmiao and Liu, Xinzhe and Shi, Jiyuan and Zhao, Yingnan and Bai, Chenjia and Li, Xuelong},
  journal={arXiv preprint arXiv:2506.08840},
  year={2025}
}

@inproceedings{vital2023,
  title={ViTAL: Vision-Based Terrain-Aware Locomotion for Legged Robots},
  author={Fawcett, R. and others},
  booktitle={IEEE/RSJ International Conference on Intelligent Robots and Systems (IROS)},
  year={2023}
}

@inproceedings{fu2023deep,
  title={Deep whole-body control: learning a unified policy for manipulation and locomotion},
  author={Fu, Zipeng and Cheng, Xuxin and Pathak, Deepak},
  booktitle={Conference on Robot Learning},
  pages={138--149},
  year={2023},
  organization={PMLR}
}

@misc{humanoid_badminton,
      title={Humanoid Whole-Body Badminton via Multi-Stage Reinforcement Learning}, 
      author={Chenhao Liu and Leyun Jiang and Yibo Wang and Kairan Yao and Jinchen Fu and Xiaoyu Ren},
      year={2025},
      eprint={2511.11218},
      archivePrefix={arXiv},
      primaryClass={cs.RO},
      url={https://arxiv.org/abs/2511.11218}, 
}

@article{him,
  title   = {Learning h-infinity locomotion control},
  author  = {Long, Junfeng and Yu, Wenye and Li, Quanyi and Wang, Zirui and Lin, Dahua and Pang, Jiangmiao},
  journal = {arXiv preprint arXiv:2404.14405},
  year    = {2024}
}

@article{anymal_parkour,
  title     = {Anymal parkour: Learning agile navigation for quadrupedal robots},
  author    = {Hoeller, David and Rudin, Nikita and Sako, Dhionis and Hutter, Marco},
  journal   = {Science Robotics},
  volume    = {9},
  number    = {88},
  pages     = {eadi7566},
  year      = {2024},
  publisher = {American Association for the Advancement of Science}
}

@misc{MoE-Loco,
  title         = {MoE-Loco: Mixture of Experts for Multitask Locomotion},
  author        = {Huang, Runhan and Zhu, Shaoting and Du, Yilun and Zhao, Hang},
  year          = {2025},
  eprint        = {2503.08564},
  archivePrefix = {arXiv},
  primaryClass  = {cs.RO},
  url           = {https://arxiv.org/abs/2503.08564}
}

@misc{omni_perception,
      title={Omni-Perception: Omnidirectional Collision Avoidance for Legged Locomotion in Dynamic Environments}, 
      author={Zifan Wang and Teli Ma and Yufei Jia and Xun Yang and Jiaming Zhou and Wenlong Ouyang and Qiang Zhang and Junwei Liang},
      year={2025},
      eprint={2505.19214},
      archivePrefix={arXiv},
      primaryClass={cs.RO},
      url={https://arxiv.org/abs/2505.19214}, 
}

@misc{perceptive_him,
      title={Learning Humanoid Locomotion with Perceptive Internal Model}, 
      author={Junfeng Long and Junli Ren and Moji Shi and Zirui Wang and Tao Huang and Ping Luo and Jiangmiao Pang},
      year={2024},
      eprint={2411.14386},
      archivePrefix={arXiv},
      primaryClass={cs.RO},
      url={https://arxiv.org/abs/2411.14386}, 
}

@misc{omni_stair,
      title={Omnidirectional Humanoid Locomotion on Stairs via Unsafe Stepping Penalty and Sparse LiDAR Elevation Mapping}, 
      author={Yuzhi Jiang and Yujun Liang and Junhao Li and Han Ding and Lijun Zhu},
      year={2026},
      eprint={2603.07928},
      archivePrefix={arXiv},
      primaryClass={cs.RO},
      url={https://arxiv.org/abs/2603.07928}, 
}

@misc{slr,
      title={SLR: Learning Quadruped Locomotion without Privileged Information}, 
      author={Shiyi Chen and Zeyu Wan and Shiyang Yan and Chun Zhang and Weiyi Zhang and Qiang Li and Debing Zhang and Fasih Ud Din Farrukh},
      year={2024},
      eprint={2406.04835},
      archivePrefix={arXiv},
      primaryClass={cs.RO},
      url={https://arxiv.org/abs/2406.04835}, 
}

@book{pattern_recognition,
  title={Pattern recognition and machine learning},
  author={Bishop, Christopher M},
  year={2006},
  publisher={Springer}
}

@misc{lggym,
      title={Learning to Walk in Minutes Using Massively Parallel Deep Reinforcement Learning}, 
      author={Nikita Rudin and David Hoeller and Philipp Reist and Marco Hutter},
      year={2022},
      eprint={2109.11978},
      archivePrefix={arXiv},
      primaryClass={cs.RO},
      url={https://arxiv.org/abs/2109.11978}, 
}

@misc{cuite,
      title={Putting the Object Back into Video Object Segmentation}, 
      author={Ho Kei Cheng and Seoung Wug Oh and Brian Price and Joon-Young Lee and Alexander Schwing},
      year={2024},
      eprint={2310.12982},
      archivePrefix={arXiv},
      primaryClass={cs.CV},
      url={https://arxiv.org/abs/2310.12982}, 
}

@misc{qwen35blog,
    title = {Qwen3.5: Accelerating Productivity with Native Multimodal Agents},
    url = {https://qwen.ai/blog?id=qwen3.5},
    author = {Qwen Team},
    month = {February},
    year = {2026}
}

@book{hartley2003multiple,
  title={Multiple View Geometry in Computer Vision},
  author={Hartley, Richard and Zisserman, Andrew},
  year={2003},
  edition={2nd},
  publisher={Cambridge University Press},
  address={Cambridge, UK},
  isbn={978-0-521-54051-3}
}

@article{schulman2017proximal,
  title={Proximal policy optimization algorithms},
  author={Schulman, John and Wolski, Filip and Dhariwal, Prafulla and Radford, Alec and Klimov, Oleg},
  journal={arXiv preprint arXiv:1707.06347},
  year={2017}
}

@inproceedings{calli2015ycb,
  title={The YCB object and model set: Towards common benchmarks for manipulation research},
  author={Calli, Berk and Singh, Arjun and Walsman, Aaron and Srinivasa, Siddhartha and Abbeel, Pieter and Dollar, Aaron M},
  booktitle={2015 international conference on advanced robotics (ICAR)},
  pages={510--517},
  year={2015},
  organization={IEEE}
}
